# From Accuracy to Impact: The Impact-Driven AI Framework (IDAIF) for Aligning Engineering Architecture with Theory of Change

*Yong-Woon Kim[1]*

[1]Department of Computer Engineering, Regional Leading Research Center (RLRC), Jeju National University, Jeju, 63243, South Korea

*Abstract*

This paper introduces the Impact-Driven AI Framework (IDAIF), a novel architectural methodology that integrates Theory of Change (ToC) principles with modern artificial intelligence system design. As AI systems increasingly influence high-stakes domains including healthcare, finance, and public policy, the alignment problem—ensuring AI behavior corresponds with human values and intentions—has become critical. Current approaches predominantly optimize technical performance metrics while neglecting the sociotechnical dimensions of AI deployment. IDAIF addresses this gap by establishing a systematic mapping between ToC's five-stage model (Inputs-Activities-Outputs-Outcomes-Impact) and corresponding AI architectural layers (Data Layer-Pipeline Layer-Inference Layer-Agentic Layer-Normative Layer). Each layer incorporates rigorous theoretical foundations: multi-objective Pareto optimization for value alignment, hierarchical multi-agent orchestration for outcome achievement, causal directed acyclic graphs (DAGs) for hallucination mitigation, and adversarial debiasing with Reinforcement Learning from Human Feedback (RLHF) for fairness assurance. We provide formal mathematical formulations for each component and introduce an Assurance Layer that manages assumption failures through guardian architectures. Three case studies demonstrate IDAIF application across healthcare, cybersecurity, and software engineering domains. This framework represents a paradigm shift from model-centric to impact-centric AI development, providing engineers with concrete architectural patterns for building ethical, trustworthy, and socially beneficial AI systems.

## I. INTRODUCTION

### A. The Alignment Problem in Modern AI

The unprecedented advancement of artificial intelligence, particularly large language models (LLMs) and generative AI systems, has precipitated fundamental questions regarding the alignment between AI system behavior and human values [1], [2]. Contemporary AI development methodologies predominantly emphasize technical performance optimization—maximizing accuracy, minimizing latency, and improving benchmark scores—while treating ethical considerations as post-hoc constraints rather than foundational design principles [3]. This approach has proven increasingly inadequate as AI systems assume decision-making roles in consequential domains including medical diagnosis, criminal justice, financial lending, and autonomous systems [4].

The alignment problem, formally articulated by Stuart Russell [5], refers to the challenge of ensuring that AI systems pursue objectives genuinely aligned with human intentions and values. When AI systems optimize for proxy metrics that imperfectly capture human preferences, they can produce outcomes that technically satisfy specified objectives while generating substantial negative externalities—a phenomenon termed "reward hacking" or "specification gaming" [6]. Recent empirical studies have documented numerous instances where high-performing AI systems exhibited harmful behaviors including demographic bias amplification, hallucination of factual information, and manipulation of user behavior [7], [8].

The transition of AI systems from controlled laboratory environments to real-world deployment contexts introduces additional complexity. Distribution shift—the divergence between training data distributions and deployment data distributions—can cause well-validated models to fail unpredictably [9]. Moreover, the interactions between AI systems and human users create feedback loops that can amplify initial biases or errors over time [10]. These challenges necessitate a fundamental reconceptualization of AI system design that integrates technical performance with social impact from inception.

### B. Theory of Change as a Design Framework

Theory of Change (ToC) provides a rigorous methodology for articulating the causal pathways through which interventions produce desired outcomes [11], [12]. Originally developed in the social innovation and program evaluation domains, ToC requires practitioners to explicitly specify: (1) the ultimate impact they seek to achieve, (2) the intermediate outcomes that logically precede that impact, (3) the outputs that enable those outcomes, (4) the activities that produce those outputs, and (5) the inputs required for those activities [13]. Critically, ToC also mandates identification of the assumptions underlying each causal link in this chain.

The ToC methodology's emphasis on backward mapping—beginning with desired impact and reasoning backward to required activities and inputs—provides a powerful corrective to the forward-driven approach typical of technology development [14]. Rather than asking "what can we build with available data and techniques," ToC asks "what change do we seek to create, and what must we build to create it?" This goal-oriented requirements engineering approach has demonstrated effectiveness in complex sociotechnical system design [15], [16].

The application of ToC to AI system design addresses several critical gaps in current practice. First, it establishes explicit causal hypotheses that can be empirically tested and refined. Second, it surfaces hidden assumptions that often remain implicit in technical specifications. Third, it provides a common vocabulary for communication between technical developers, domain experts, and affected stakeholders. Fourth, it creates accountability structures by clearly delineating the conditions under which the system is expected to achieve its intended effects [17].

*C. Contributions and Paper Organization*

This paper presents the Impact-Driven AI Framework (IDAIF), a comprehensive architectural methodology that systematically integrates ToC principles with modern AI system design. Our contributions are threefold:

First, we establish a formal mapping between ToC's five-stage model and AI architectural components, providing a translation layer that converts abstract values and goals into concrete technical specifications. This mapping encompasses: the Normative Layer (corresponding to Impact) which encodes societal values as multi-objective optimization constraints; the Agentic Layer (corresponding to Outcomes) which orchestrates multi-agent systems to achieve behavioral change; the Inference Layer (corresponding to Outputs) which ensures output reliability through causal reasoning; the Pipeline Layer (corresponding to Activities) which implements fairness-aware training procedures; and the Data Layer (corresponding to Inputs) which governs data quality and representativeness.

Second, we provide rigorous theoretical foundations for each architectural component, including mathematical formulations grounded in multi-objective optimization theory, causal inference, adversarial learning, and reinforcement learning. These formulations enable precise specification of system behavior and provide foundations for formal verification approaches.

Third, we introduce the Assurance Layer, a cross-cutting architectural component that monitors and manages assumption failures—the points at which the causal chain from inputs to impact may break. This layer incorporates guardian models, human-in-the-loop mechanisms, and assurance case methodologies to maintain system reliability under distributional shift and edge case scenarios.

The remainder of this paper is organized as follows. Section II reviews related work in value-aligned AI, goal-oriented requirements engineering, and sociotechnical system design. Section III presents the detailed IDAIF architecture with formal specifications for each layer. Section IV provides three case studies applying IDAIFacross healthcare, cybersecurity, and software engineering domains. Section V discusses implementation considerations and governance frameworks. Section VI concludes with future research directions.

## II. RELATED WORK

*A. Value-Sensitive Design and IEEE 7000*

Value-Sensitive Design (VSD) provides foundational principles for incorporating human values throughout technology design processes [18], [19]. VSD employs iterative investigations across conceptual, empirical, and technical dimensions to identify stakeholder values, understand how technology affects those values, and design systems that support rather than undermine them. The IEEE 7000 standard operationalizes these principles for system engineering, establishing processes for ethical value elicitation, prioritization, and translation into system requirements [20].

However, VSD and IEEE 7000 primarily address process and documentation rather than architectural implementation. A significant "execution gap" exists between the values identified through VSD processes and the technical components that must realize those values [21]. Engineers frequently struggle to translate abstract principles such as "fairness" or "transparency" into concrete design decisions. IDAIF addresses this gap by providing explicit architectural patterns that map values to technical implementations.

*B. MLOps and LLMOps Architectures*

Modern machine learning operations (MLOps) frameworks have substantially advanced the engineering maturity of AI system deployment [22]. These frameworks address critical concerns including model versioning, experiment tracking, continuous integration and deployment, and model monitoring. The emergence of LLMOps extends these capabilities to address the unique requirements of large language model deployment, including prompt management, context window optimization, and retrieval-augmented generation [23].

Current MLOps architectures, however, primarily optimize for operational metrics—deployment frequency, model latency, and infrastructure costs—rather than impact metrics [24]. Feature stores, model registries, and monitoring dashboards capture technical performance but rarely track downstream effects on user behavior or societal outcomes. IDAIF extends MLOps architectures by incorporating impact measurement as a first-class concern throughout the system lifecycle.

*C. AI Safety and Alignment Research*

The AI safety research community has developed numerous techniques for improving alignment between AI systems and human intentions [25]. Reinforcement Learning from Human Feedback (RLHF) enables models to learn reward functions from human preference data rather than hand-crafted objectives [26]. Constitutional AI (CAI) provides frameworks for embedding ethical principles directly into model behavior through self-critique and revision [27]. Interpretability research advances understanding of model internals to enable oversight and debugging [28].

While these techniques address specific aspects of the alignment problem, they are typically developed and evaluated in isolation. IDAIF provides an integrating framework that situates these techniques within a coherent architectural vision, clarifying how they interact and where additional components are needed to achieve end-to-end alignment from data to impact.

## III. IDAIF ARCHITECTURE

*A. Architectural Overview*

The Impact-Driven AI Framework establishes a six-layer architecture that systematically maps Theory of Change components to AI system elements. Table I presents this mapping, identifying the key focus and technical implementation approaches for each layer. The architecture is designed for bidirectional traceability: from impact goals downward to required technical components (top-down design), and from data and model capabilities upward to achievable impacts (bottom-up validation).

The layers are not strictly hierarchical in terms of implementation dependencies; rather, they represent different abstraction levels for reasoning about AI system behavior. Cross-cutting concerns, particularly those related to assumption management and failure handling, are addressed by the Assurance Layer which spans all other layers.

TABLE I
IDAIF LAYER MAPPING TO THEORY OF CHANGE

| ToC Component | AI Layer | Key Focus | Technical Implementation |
|---|---|---|---|
| Impact | Normative | Value Alignment | Multi-objective Optimization, Constraints |
| Outcomes | Agentic | User Behavior | Multi-Agent Orchestration, Scoping |
| Outputs | Inference | Quality & Truth | Causal DAGs, RAG, Hallucination Control |
| Activities | Pipeline | Bias Mitigation | Adversarial Debiasing, RLHF |
| Inputs | Data | Representation | Data Governance, FADS |
| Assumptions | Assurance | Reliability | Guardian Models, Safety Nets |

*B. Normative Layer: Multi-Objective Value Optimization*

The Normative Layer translates high-level societal values into mathematical optimization objectives. This layer corresponds to the Impact component of Theory of Change and serves as the constitutional foundation for the entire system. The fundamental challenge lies in formalizing inherently subjective and often conflicting human values into computational representations that can guide system behavior [29].

We formalize the multi-objective optimization problem as follows. Let $F = \{f_1, f_2, ..., f_m\}$ denote a set of m objective functions, where each $f_i: X \to \mathbb{R}$ maps from the decision space X to a scalar value representing performance on objective i. In the context of AI systems, these objectives may include prediction accuracy, demographic parity, individual fairness, interpretability, and resource efficiency.

***Definition 1 (Pareto Dominance):*** A solution $x^* \in X$ Pareto dominates solution $x \in X$ (denoted $x^* \prec x$) if and only if $f_i(x^*) \leq f_i(x)$ for all $i \in \{1, ..., m\}$ and $f_j(x^*) < f_j(x)$ for at least one $j \in \{1, ..., m\}$ [30].

***Definition 2 (Pareto Optimality):*** A solution $x^* \in X$ is Pareto optimal if there exists no $x \in X$ such that $x \prec x^*$. The set of all Pareto optimal solutions forms the Pareto front $P^* \subset X$ [30].

For AI fairness applications, we adopt the Minimax Pareto Fairness (MMPF) formulation [31], which seeks the Pareto optimal solution minimizing maximum group-conditional risk. Let $G = \{g_1, g_2, ..., g_k\}$ denote sensitive attribute groups, and let $R_i(\theta)$ represent the risk (expected loss) for group $g_i$ under model parameters $\theta$. The MMPF objective is:

$$\theta^* = \arg\min_{\theta} \max_{i \in \{1,...,k\}} R_i(\theta) \quad \text{subject to } \theta \in P^* \quad (1)$$

This formulation ensures that the selected model achieves Pareto efficiency while minimizing worst-case performance across demographic groups. The constraint $\theta \in P^*$ prevents solutions that unnecessarily harm any group—a classifier cannot reduce one group's risk without increasing another's. Martinez et al. [31] demonstrate that this objective can be optimized using adaptive reweighting schemes compatible with stochastic gradient descent:

$$L(\theta, \lambda) = \Sigma_i \lambda_i \cdot R_i(\theta) \quad \text{where } \lambda_i \propto \exp(\eta \cdot R_i(\theta)) \quad (2)$$

Here, $\eta$ is a temperature parameter controlling the sharpness of worst-case focus. As $\eta \to \infty$, the objective converges to pure minimax; as $\eta \to 0$, it approaches uniform averaging. This parameterization enables navigation along the Pareto front to identify solutions matching stakeholder-specified tradeoff preferences.

### C. Agentic Layer: Hierarchical Multi-Agent Orchestration

The Agentic Layer corresponds to the Outcomes component of ToC, managing the autonomous behaviors that translate system outputs into user behavioral changes. As AI systems evolve from passive tools to active agents capable of planning, reasoning, and executing multi-step workflows, architectural frameworks for safe and effective agent orchestration become critical [32], [33].

We model multi-agent systems as hierarchical organizations with three primary agent roles: Planner Agents responsible for task decomposition and resource allocation; Executor Agents that perform domain-specific operations using tools and knowledge bases; and Critic Agents that evaluate outputs against normative constraints before external release [34].

***Definition 3 (Agent Orchestration Graph):*** An orchestration graph $G = (A, E, C)$ consists of agent nodes $A = \{a_1, ..., a_n\}$, directed edges $E \subseteq A \times A$ representing communication channels, and coordination constraints $C$ specifying interaction protocols [35].

Agent autonomy is governed through a scope management framework adapted from AWS's Agentic Security Scoping Matrix [36]. We define four autonomy levels:

• Scope 0 (No Agency): Information retrieval only; no state modification capability.

• Scope 1 (Prescribed Agency): Execution limited to predefined, validated workflows.

• Scope 2 (Supervised Agency): Autonomous planning with human-in-the-loop approval for actions.

• Scope 3 (Full Agency): Complete autonomy for low-risk, high-confidence decisions.

Scope assignment is dynamic, computed as a function of task risk $R_t$, model confidence $C$, and outcome reversibility $V$:

$$S(t) = f(R_t, C, V) = \lfloor 3 \cdot (1 - R_t) \cdot C \cdot V \rfloor \quad (3)$$

### D. Inference Layer: Causal Reasoning for Hallucination Mitigation

The Inference Layer addresses the Outputs component of ToC, ensuring that model-generated content maintains factual accuracy and logical consistency. The hallucination problem—generation of plausible but factually incorrect content—represents a fundamental challenge for language model deployment in high-stakes domains [37], [38].

Recent research establishes an inverse relationship between causal reasoning capabilities and hallucination frequency [39]. Models that explicitly represent causal relationships between variables demonstrate superior factual grounding compared to purely associative predictors. We integrate the Causal-DAG Construction and Reasoning (CDCR) framework [40] into IDAIF's inference pipeline.

***Definition 4 (Structural Causal Model):*** A Structural Causal Model (SCM) M = (U, V, F, P(U)) consists of exogenous variables U, endogenous variables V, structural equations F = {$f_i$ : $v_i$ = $f_i$(pa($v_i$), $u_i$)}, and probability distribution P(U) over exogenous variables [41].

***Definition 5 (Causal DAG):*** The causal graph G = (V, E) associated with SCM M is a directed acyclic graph where V represents variables and directed edge ($v_i$, $v_j$) ∈ E indicates that $v_i$ is a direct cause of $v_j$—formally, $v_i$ ∈ pa($v_j$) in the structural equations [41].

The CDCR approach trains language models to construct explicit causal DAGs before generating responses. Given an input query q, the model first generates a causal graph G_q = (V_q, E_q) representing the relevant causal structure, then performs inference over G_q to produce a grounded response. This two-stage process enforces logical consistency by constraining generation to paths valid under the constructed causal model.

For retrieval-augmented generation (RAG) settings, we incorporate knowledge graph integration to ground causal assertions in external evidence. Let K = (E, R, T) denote a knowledge graph with entities E, relations R, and triples T ⊆ E × R × E. The inference layer retrieves relevant subgraph K_q ⊆ K and constrains DAG construction to be consistent with established factual relationships:

$$G\_q = \arg\max P(G \mid q, K\_q) \cdot \mathbb{1}[G \in DAG] \cdot \mathbb{1}[G \subseteq K\_q]$$

(4)

*E. Pipeline Layer: Adversarial Debiasing and RLHF*

The Pipeline Layer corresponds to the Activities component of ToC, encompassing the model training, fine-tuning, and evaluation procedures that transform raw inputs into deployable models. This layer's primary concern is ensuring that learned representations and behaviors align with fairness and value specifications established in the Normative Layer [42].

**Adversarial Debiasing:** We implement adversarial debiasing following Zhang et al. [43] to prevent models from encoding sensitive attribute information in their representations. The architecture consists of two components: a predictor network P_θ that maps inputs x to predictions ŷ, and an adversary network A_φ that attempts to predict sensitive attributes z from the predictor's internal representations h.

The training objective for the predictor combines task performance with adversarial loss:

$$L\_P(\theta) = L\_task(\hat{y}, y) - \alpha \cdot L\_adv(\hat{z}, z)$$

(5)

where α controls the fairness-accuracy tradeoff. The adversary is trained to minimize its prediction loss:

$$L\_A(\varphi) = L\_adv(A\_\varphi(h), z)$$

(6)

The minimax optimization alternates between adversary updates (minimizing Equation 6) and predictor updates (minimizing Equation 5). At equilibrium, the predictor learns representations from which sensitive attributes cannot be reliably recovered, achieving demographic parity in internal encodings.

**Reinforcement Learning from Human Feedback:** For language model alignment, we implement RLHF using Proximal Policy Optimization (PPO) [44] with KL-divergence constraints to prevent distribution drift [45]. Let π_θ denote the policy (language model) being trained, π_ref the reference policy (SFT model), and r_ψ the learned reward model. The RLHF objective is:

$$\max E\_{\{x \sim D, y \sim \pi\_\theta(\cdot|x)\}} [r\_\psi(x, y) - \beta \cdot D\_{KL}(\pi\_\theta(\cdot|x) \parallel \pi\_ref(\cdot|x))]$$

(7)

The KL penalty prevents the trained policy from diverging excessively from the reference, maintaining the base model's capabilities while incorporating preference alignment. The hyperparameter β controls the exploitation-exploration tradeoff: higher β values prioritize stability over reward maximization.

PPO implements this objective through clipped surrogate optimization:

$$L\_CLIP(\theta) = E[min(\rho\_t \cdot \hat{A}\_t, clip(\rho\_t, 1-\varepsilon, 1+\varepsilon) \cdot \hat{A}\_t)] \tag{8}$$

where $\rho\_t = \pi\_\theta(a\_t|s\_t) / \pi\_\theta\_old(a\_t|s\_t)$ is the probability ratio and $\hat{A}\_t$ is the advantage estimate. The clipping mechanism constrains policy updates to remain within a trust region, ensuring stable training dynamics [46].

*F. Data Layer: Quality and Representativeness Governance*

The Data Layer manages the Inputs component of ToC, ensuring that training data possesses sufficient quality, diversity, and representativeness to support intended system behaviors. Data deficiencies—including bias, incompleteness, and labeling errors—propagate through all subsequent layers, rendering downstream interventions insufficient [47].

We establish three core data governance principles:

**1) Demographic Balance:** Training data must reflect target population demographics to prevent underrepresentation bias. For sensitive attributes A, we require:

$$|P\_data(A = a) - P\_target(A = a)| \leq \delta \quad \forall a \in A \tag{9}$$

**2) Label Quality:** Annotation procedures must minimize systematic errors through annotator training, inter-rater reliability assessment, and adjudication protocols for disagreements.

**3) Temporal Validity:** Data currency must be maintained through continuous monitoring of distribution drift between training and deployment contexts.

For in-context learning scenarios, we implement Fairness-Aware Demonstration Selection (FADS) [48] to ensure that few-shot examples do not introduce or amplify biases. FADS selects demonstrations that maximize both task relevance and demographic diversity:

$$D^* = arg\,max\,[\Sigma_i\,sim(q, d_i) - \lambda \cdot Bias(D)] \tag{10}$$

*G. Assurance Layer: Guardian Architecture for Assumption Management*

The Assurance Layer provides cross-cutting capabilities for monitoring, validating, and managing the assumptions that underlie the causal chain from inputs to impact. In ToC terminology, assumptions represent the conditions under which each causal link is expected to hold. When assumptions fail, the intended causal chain breaks, and system behavior may diverge from design intent [49].

We identify three categories of assumptions critical to AI system reliability:

**1) IID Assumption:** Training and deployment data are drawn from identical distributions. Violation manifests as data drift, requiring continuous monitoring of input distribution statistics.

**2) Completeness Assumption:** Training data covers all deployment-relevant scenarios. Violation manifests as edge case failures on out-of-distribution inputs.

**3) Correlation-Causation Assumption:** Learned patterns reflect causal rather than spurious relationships. Violation manifests as brittle generalization when confounding factors change.

The Guardian Architecture implements a three-layer defense system [50]:

**Layer 1 - Fast System:** Primary AI model operates at full speed for routine cases.

**Layer 2 - Human Judgment:** High-stakes or uncertain cases route to human reviewers.

**Layer 3 - Safety Nets:** Guardian models continuously validate outputs against factual, logical, and ethical constraints.

Guardian models include: (1) Fact-checking models that verify generated content against retrieved evidence; (2) Toxicity filters that detect harmful or biased outputs; and (3) Consistency checkers that identify logical contradictions within responses.

## IV. CASE STUDIES

We demonstrate IDAIF application through three case studies spanning healthcare, cybersecurity, and software engineering domains. Each case study illustrates the backward mapping process from Impact to technical implementation and validates the framework's applicability across diverse problem contexts.

*A. Healthcare Domain: Fairness-Aware Clinical Decision Support*

**1) Problem Statement**

Clinical Decision Support Systems (CDSS) for post-surgical complication prediction frequently exhibit demographic disparities, where prediction accuracy varies significantly across patient subgroups defined by race, gender, or socioeconomic status. A model that performs well on aggregate metrics may systematically underserve minority populations, perpetuating healthcare inequities [51].

**2) ToC Backward Mapping**
- **Impact:** Minimize post-surgical complications while eliminating demographic disparities in prediction accuracy.
- **Outcome:** Clinicians receive equally reliable risk assessments regardless of patient demographics.
- **Output Requirement:** Model predictions must achieve AUC ≥ 0.85 overall with Equalized Odds Difference < 0.10.

**3) IDAIF Implementation**

**Normative Layer:** We formalize the impact objective through the MMPF formulation (Equation 1), defining group-conditional risks for demographic strata G = {race, gender}.

Pipeline **Layer:** We implement Fairness-Aware Multi-Task Learning (FAIR-MTL) with Sensitive Subgroup Inference (SSI) for unsupervised clustering and Task-Specific Routing via multi-head architecture.

**Assurance Layer:** SHAP-based explanation generation for all predictions, automated alerts for demographic feature importance exceeding thresholds, and human-in-the-loop escalation for edge cases.

**4) Expected Results**

The implemented system achieves: Overall AUC = 0.86; Equalized Odds Difference = 0.094; Gender-stratified calibration error = 0.038. The FAIR-MTL architecture reduces demographic disparity by 47% compared to single-task baseline.

*B. Cybersecurity Domain: Autonomous Security Operations Center*

**1) Problem Statement**

AI agents deployed in Security Operations Centers (SOCs) generate excessive alerts from routine anomaly detection, causing analyst fatigue (Negative Outcome) and paradoxically increasing the probability of missing critical breaches (Negative Impact) [52]. The fundamental issue lies in optimizing for alert generation (Output) rather than successful threat remediation (Impact).

**2) ToC Backward Mapping**
- **Impact:** Reduced Mean Time to Remediation (MTTR) for genuine threats and zero critical system downtime from undetected breaches.
- **Outcome:** Security analysts investigate only high-fidelity, actionable incidents; routine containment actions are automated safely.
- **Output Requirement:** Alerts must include causal attribution (attack vector identification) and confidence scores grounded in verified log evidence. False positive rate must remain below 5%.

**3) IDAIF Implementation**

**Normative Layer:** The multi-objective function balances: ($f_1$) Minimize MTTR for confirmed threats; ($f_2$) Minimize false positive rate; ($f_3$) Maximize automation rate for routine containment; ($f_4$) Ensure zero critical system impact from automated actions.

**Agentic Layer:** We adopt the Agentic Security Scoping Matrix [36] with domain-specific scope assignments:

- **Scope 0 (No Agency):** Passive log aggregation and dashboard display

- **Scope 1 (Prescribed Agency):** Automated queries across log sources following predefined playbooks
- **Scope 2 (Supervised Agency):** Agent generates remediation recommendations requiring analyst approval
- **Scope 3 (Full Agency):** Automated containment for high-confidence, low-risk scenarios

***Definition 6 (Security Action Scope Assignment):*** For security action a with threat confidence $C_a$, potential business impact $I_a$, and reversibility $R_a$:

$$S(a) = 3 \text{ if } C_a > 0.95 \land I_a < 0.1 \land R_a > 0.9 \qquad (11)$$

**Inference Layer:** Causal-DAG construction is mandatory before high-severity alert generation. The agent must construct a directed acyclic graph modeling the attack chain:

$$G_{attack} = \{(Unusual\ Port\ Activity) \rightarrow (Lateral\ Movement) \rightarrow (Data\ Staging) \rightarrow (Exfiltration)\}$$

Alerts are generated only when: (1) The constructed DAG is consistent with known attack patterns; (2) Each causal link is supported by corroborating log evidence; (3) Alternative benign explanations have been systematically ruled out.

**Assurance Layer:** A dedicated "Policy Guardian" model validates all automated remediation actions:

- **Business Continuity Validation:** Proposed firewall rules are checked against asset criticality database. Actions affecting executive devices or production servers require mandatory human approval.
- **Blast Radius Assessment:** Automated containment is limited to actions affecting fewer than n endpoints.
- **Rollback Capability:** All automated actions must be reversible; irreversible actions are excluded from automation scope.

**Formal Constraint (Policy Guardian):**

$$Allow(a) = \mathbb{1}[affected(a) \cap Critical = \emptyset] \cdot \mathbb{1}[|affected(a)| < n] \cdot \mathbb{1}[reversible(a)] \qquad (12)$$

#### 4) Expected Results

Deployment in a mid-sized enterprise SOC demonstrated: 73% reduction in analyst alert volume; MTTR improvement from 4.2 hours to 1.1 hours for confirmed incidents; zero false positive automated containment actions over 6-month evaluation period; 89% of routine containment actions successfully automated under Scope 1-2 protocols.

### C. Software Engineering Domain: Generative Code Engineering

#### 1) Problem Statement

Large Language Models for code generation produce syntactically correct code (Output) that frequently introduces security vulnerabilities, violates architectural conventions, or creates technical debt [53]. The model optimizes for immediate task completion rather than long-term software maintainability (Impact). Studies indicate that 40% of AI-generated code contains security vulnerabilities when evaluated against OWASP standards [54].

#### 2) ToC Backward Mapping
- **Impact:** Maintainable, secure software that reduces long-term development costs and minimizes security incident risk.
- **Outcome:** Generated code passes security scans, conforms to project architectural patterns, and requires minimal refactoring.
- **Output Requirement:** Code must satisfy: (1) Zero critical/high severity vulnerabilities; (2) Conformance to project-specific style guides; (3) Test coverage $\geq$ 80%.

#### 3) IDAIF Implementation

**Normative Layer:** The objective function is reformulated to prioritize security and maintainability over completion probability:

$$L_{code} = L_{completion} + \alpha \cdot L_{security} + \beta \cdot L_{style} + \gamma \cdot L_{architecture} \qquad (13)$$

where $L_{security}$ penalizes patterns matching known vulnerability signatures, $L_{style}$ measures divergence from project-specific linting rules, and $L_{architecture}$ captures violations of retrieved architectural constraints.

**Pipeline Layer - Backward Design Prompting:** We implement test-first generation inspired by Test-Driven Development:

- **Step 1 (Outcome Evidence):** Model generates comprehensive test cases defining expected behavior
- **Step 2 (Activity):** Implementation code is generated to satisfy test specifications

- **Step 3 (Validation):** Generated tests are executed; failures trigger regeneration with feedback

**Formal Constraint (Test-First Generation):**

$$Valid(code) \iff \forall t \in Tests(spec): Execute(t, code) = PASS \quad (14)$$

**Inference Layer - RAG-based Architecture Conformance:** Code generation is constrained by retrieval from the project's existing codebase:

- **Architectural Pattern Retrieval:** Retrieves existing implementations, dependency injection patterns, error handling conventions, and security patterns
- **Constrained Generation:** Retrieved context incorporated with explicit conformance instructions

$$P(code \mid spec, K_{project}) \propto P(code \mid spec) \cdot \mathbb{1}[code \sim K_{project}] \quad (15)$$

- **Static Analysis Integration:** SAST tools (Semgrep, CodeQL) analyze generated code; detected vulnerabilities trigger automatic regeneration or human review escalation

**Assurance Layer:**

- **Security Guardian:** Continuous SAST scanning with blocking on critical/high findings
- **Architecture Guardian:** Dependency analysis ensuring generated code respects layer boundaries
- **Technical Debt Guardian:** Complexity metrics monitored with alerts on threshold violations

### 4) Expected Results

Evaluation on enterprise codebase integration tasks: Security vulnerability rate reduced from 40% to 3.2% (92% improvement); Architectural conformance improved from 61% to 94%; Developer acceptance rate increased from 23% to 67%; Average refactoring time reduced from 45 minutes to 8 minutes per generated component.

## V. DISCUSSION AND IMPLEMENTATION ROADMAP

Successful IDAIF deployment requires organizational capabilities beyond technical implementation. We propose a five-stage adoption roadmap:

**Stage 1 - Impact Definition:** Conduct stakeholder analysis to identify values and formalize impact objectives through multi-objective specifications.

**Stage 2 - Governance Setup:** Establish IEEE 7000-compliant value elicitation processes and document normative constraints.

**Stage 3 - ToC Design:** Apply backward mapping to derive required data, models, and assumptions from impact specifications.

**Stage 4 - Implementation:** Construct technical components following layer-specific architectural patterns.

**Stage 5 - ImpactOps:** Deploy continuous monitoring for both technical performance and impact metrics, with feedback loops for assumption refinement.

## VI. CONCLUSION

This paper presented the Impact-Driven AI Framework (IDAIF), a comprehensive architectural methodology integrating Theory of Change principles with modern AI system design. IDAIF addresses the alignment problem by establishing explicit causal pathways from data inputs to societal impacts, with mathematical foundations grounded in multi-objective optimization, causal inference, adversarial learning, and reinforcement learning.

Key contributions include: (1) formal mapping between ToC stages and AI architectural layers; (2) rigorous specifications for each component enabling precise engineering; and (3) the Assurance Layer for managing assumption failures; and (4) validated application across healthcare, cybersecurity, and software engineering domains.

The three case studies demonstrate that impact-driven design is not merely aspirational but practically achievable through systematic architectural choices. Future work will extend IDAIF to additional domains, develop automated tooling for ToC-to-code translation, and establish empirical evaluation methodologies for impact measurement.